\documentclass[10pt,twocolumn]{article} 
\usepackage{titling} 

\usepackage{iccv}
\usepackage{times}
\usepackage{epsfig}
\usepackage{graphicx}
\usepackage{amsmath}
\usepackage{amssymb}
\usepackage{multirow}

\def \ie {\emph{i.e.}\xspace}
\def \eg {\emph{e.g.}\xspace}

\def \etal {\emph{et al.}\xspace}

\def \bset {\mathcal B}

\def \aset {\mathcal A}

\newcommand{\para}[1]{\noindent \textbf{#1}}

\DeclareMathAlphabet{\mathpzc}{T1}{pzc}{m}{n}

\usepackage{tabularx} 
\usepackage{caption}

\newcommand{\pretabspace}{\vspace{-2mm}}
\newcommand{\posttabspace}{\vspace{-3mm}}
\newcommand{\prefigspace}{\vspace{-4mm}}
\newcommand{\postfigspace}{\vspace{-4mm}}
\newcolumntype{C}[1]{>{\centering\let\newline\\\arraybackslash\hspace{0pt}}m{#1}} 
\makeatletter

\usepackage[pagebackref=true,breaklinks=true,letterpaper=true,colorlinks,bookmarks=false]{hyperref}

\iccvfinalcopy 

\def\iccvPaperID{1318} 

\ificcvfinal\fi

\begin{document}

\title{Weakly Supervised Object Localization Using Things and Stuff Transfer}

\author{
Miaojing Shi$^{1,2}$\\
\\
\and
Holger Caesar$^1$\\
$^1$University of Edinburgh~~~$^2$Tencent Youtu Lab\\
{\tt\small name.surname@ed.ac.uk}
\and
Vittorio Ferrari$^1$\\
\\
}

\maketitle
\thispagestyle{empty} 

\begin{abstract}

We propose to help weakly supervised object localization for classes where location annotations are not available, by transferring things and stuff knowledge from a source set with available annotations.
The source and target classes might share similar appearance (\eg bear fur is similar to cat fur) or appear against similar background (\eg horse and sheep appear against grass).
To exploit this, we acquire three types of knowledge from the source set:
a segmentation model trained on both thing and stuff classes;
similarity relations between target and source classes;
and co-occurrence relations between thing and stuff classes in the source.
The segmentation model is used to generate thing and stuff segmentation maps on a target image, while the class similarity and co-occurrence knowledge help refining them.
We then incorporate these maps as new cues into a multiple instance learning framework (MIL), propagating the transferred knowledge from the pixel level to the object proposal level.
In extensive experiments, we conduct our transfer from the PASCAL Context dataset (source) to the ILSVRC, COCO and PASCAL VOC 2007 datasets (targets).
We evaluate our transfer across widely different thing classes, including some that are not similar in appearance, but appear against similar background.
The results demonstrate significant improvement over standard MIL, and we outperform the state-of-the-art in the transfer setting.

\end{abstract}

\section{Introduction}
\label{sec:introduction}

\begin{figure*}[t]
	\centering
	\includegraphics[width=0.99\textwidth]{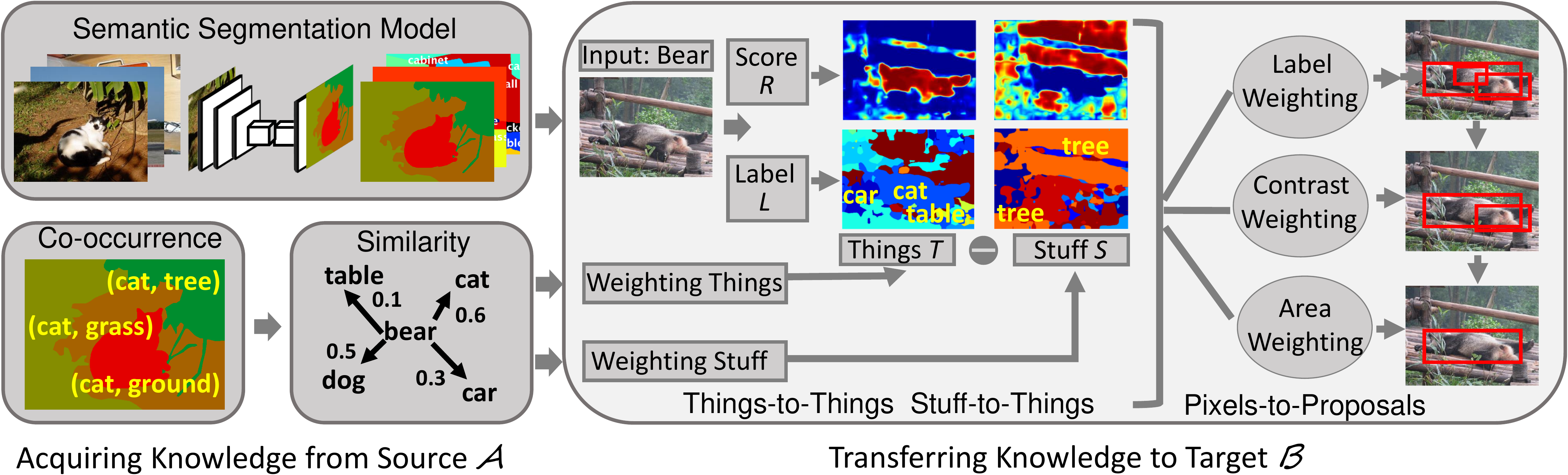}
         \prefigspace
         \vspace{2mm}
	\caption{\small An overview of our things and stuff transfer (TST) method. We acquire the 1) segmentation model, 2) co-occurrence relation and 3) similarity relation from the source $\aset$ and transfer them to the target $\bset$.
We use the segmentation model to generate two maps: thing ($T$) and stuff ($S$) maps; each of them contains one score ($R$) map and one label ($L$) map.
The knowledge of class similarity and co-occurrence is specifically transferred as weighting functions to the thing and stuff label maps.
Based on the transferred knowledge, we propose three scoring schemes (label weighting, contrast weighting, and area weighting) to propagate the information from pixels to proposals.
The rightmost image column illustrates some highly ranked proposals in the image by gradually adopting the three schemes.}
	\label{Fig:Overview}
	\postfigspace
\end{figure*}

The goal of object class detection is to place a tight bounding box on every instance of an object class. Given an input image, recent object
detectors~\cite{girshick15cvpr,girshick15iccv,cinbis14cvpr,cinbis15pami} first extract object proposals ~\cite{alexe10cvpr,uijlings13ijcv,dollar14eccv} and then score them with a classifier to determine their probabilities of containing an instance of the class. Manually annotated bounding boxes are typically required for training (full supervision).

Annotating bounding boxes is tedious and time-consuming. In order to reduce the annotation cost, many previous works learn the detector in a weakly supervised setting~\cite{bilen14bmvc,bilen15cvpr,cinbis15pami,deselaers10eccv,russakovsky12eccv,siva11iccv,song14icml,song14nips,shi16eccv},
\ie given a set of images known to contain instances of a certain object class, but without their locations. This weakly supervised object localization (WSOL) bypasses the need for bounding box annotation and substantially reduces annotation time.

Despite the low annotation cost, the performance of WSOL is considerably lower than that of full supervision. To improve WSOL, various advanced cues can be added, \eg objectness~\cite{deselaers10eccv,alexe12pami,cinbis15pami,siva11iccv,tang14cvpr, shi16eccv}, which gives an estimation of how likely a proposal contains an object;
co-occurrence among multiple classes in the same training images~\cite{shi12bmvc};
object size estimates based on an auxiliary dataset with size annotations~\cite{shi16eccv};
and appearance models transferred from object classes with bounding box annotations to new object classes~\cite{guillaumin12cvpr,hoffman14nips,rochan15cvpr}.

There are two types of classes that can be transferred from a source set with manually annotated locations:
things (objects) and stuff (materials and backgrounds).
Things have a specific spatial extent and shape (\eg helicopter, cow, car), while stuff does not (\eg sky, grass, road).
Current transfer works mostly focus on transferring appearance models among similar thing classes~\cite{guillaumin12cvpr,rochan15cvpr,hoffman14nips} (\emph{things-to-things}).
In contrast, using stuff to find things~\cite{heitz08eccv,lee11cvpr} is largely unexplored, particularly in the WSOL setting (\emph{stuff-to-things}).

In this paper, we transfer a fully supervised segmentation model from the source set to help WSOL on the target set.
We introduce several schemes to conduct the transfer of both things and stuff knowledge, guided by the similarity between classes.
Particularly, we transfer the co-occurrence knowledge between thing and stuff classes in the source via a second order scheme to thing classes in the target.
We propagate the transferred knowledge from the pixel level to the object proposal level and inject it as a new cue into a multiple instance learning framework (MIL).

In extensive experiments, we show that our method:
(1) improves over a standard MIL baseline on three datasets: ILSVRC~\cite{russakovsky15ijcv}, COCO~\cite{lin14eccv}, PASCAL VOC 2007~\cite{everingham10ijcv};
(2) outperforms the things-to-things transfer method~\cite{rochan15cvpr} and the state-of-the-art WSOL methods~\cite{bilen16cvpr,cinbis15pami,wang15tip} on VOC 2007;
(3) outperforms another things-to-things transfer method (LSDA~\cite{hoffman14nips}) on ILSVRC. 
\section{Related Work}\label{sec:relatedwork}

\para{Weakly supervised object localization.}
In WSOL the training images are known to contain instances of a certain object class but their locations are unknown. The task is both to localize the objects in the training images and to learn a detector for the class.

Due to the use of strong CNN features~\cite{girshick14cvpr,krizhevsky12nips}, recent works on WSOL~\cite{bilen15cvpr,cinbis15pami,song14icml,wang15tip,bilen16cvpr, shi16eccv} have shown remarkable progress.
Moreover, researchers also tried to incorporate various advanced cues into the WSOL process,
\eg objectness~\cite{cinbis15pami,deselaers10eccv,siva11iccv,tang14cvpr,alexe12pami}, object size~\cite{shi16eccv}, co-occurrence~\cite{shi12bmvc} among classes,
and transferring appearance models of the source thing classes to help localize similar target thing classes~\cite{guillaumin12cvpr,rochan15cvpr,hoffman14nips}.
This paper introduces a new cue called things and stuff transfer (TST),
which learns a semantic segmentation model from the source on both things and stuff annotations and transfers its knowledge to help localize the target thing class.

\para{Transfer learning.}
The goal of transfer learning is to improve the learning of a target task by leveraging knowledge from a source task~\cite{pan10tkde}.
It is intensively studied in image classification, segmentation and object detection~\cite{aytar2011iccv,aytar12bmvc,tommasi2010cvpr,lampert:cvpr09,rohrbach10cvpr,ott2011cvpr,stark:iccv09}.
Many methods use the parameters of the source classifiers as priors for the target model~\cite{aytar2011iccv,aytar12bmvc,tommasi2010cvpr}.
Other works~\cite{lampert:cvpr09,rohrbach10cvpr} transfer knowledge through an intermediate attribute layer, which captures visual qualities shared by many object classes (\eg ``striped", ``yellow").
A third family of works transfer object parts between classes~\cite{aytar12bmvc,ott2011cvpr,stark:iccv09}, \eg wheels between cars and bicycles.

In this work we are interested in the task where we have the location annotations in the source and transfer them to help learn the classes in the target~\cite{shi12bmvc,kuettel12eccv,rochan15cvpr,guillaumin12cvpr,lee11cvpr,heitz08eccv}.
We categorize the transfer into two types:
1) \emph{Things-to-things.} Guillaumin \etal~\cite{guillaumin12cvpr} transferred spatial location, appearance, and context information from the source thing classes to localize the things in the target;
Shi \etal~\cite{shi12bmvc} and Rochan \etal~\cite{rochan15cvpr} follow a similar spirit to~\cite{guillaumin12cvpr};
while Kuettel \etal~\cite{kuettel12eccv} instead transferred segmentation masks.
2) \emph{Stuff-to-things.} Heitz \etal~\cite{heitz08eccv} proposed a context model to utilize stuff regions to find things, in a fully supervised setting for the target objects;
Lee \etal~\cite{lee11cvpr} also made use of stuff annotations in the source to discover things in the target, in an unsupervised setting.

Our work offers several new elements over these:
(1) we encode the transfer as a combination of both \emph{things-to-things} and \emph{stuff-to-things};
(2) we propose a model to propagate the transferred knowledge from the pixel level to the proposal level;
(3) we introduce a second order transfer, \ie \emph{stuff-to-things-to-things}. 
\section{Overview of our method}\label{sec:method}

In this section we define the notations and introduce our method on a high level, providing some details for each part.

\para{Notations.}
We have a source set $\aset$ and a target set $\bset$.
We have every image pixelwise annotated for both stuff and things in $\aset$; whereas we have only image level labels for images in $\bset$.
We denote  by $\aset^T$ the set of thing classes in $\aset$, and $a^t$ an individual thing class; analogue we have $\aset^S$ and $a^s$ for stuff classes in $\aset$ and $\bset^T$ and $b^t$ for thing classes in $\bset$.
Note that there are no stuff classes in $\bset$, as datasets labeled only by thing classes are more common in practice (\eg PASCAL VOC~\cite{everingham15ijcv}, ImageNet~\cite{russakovsky15ijcv}, COCO~\cite{lin14eccv}).

\para{Method overview.}
Our goal is to conduct WSOL on $\bset$, where the training images are known to contain instances of a certain object class but their locations are unknown.
A standard WSOL approach, \eg MIL, treats images as bags of object proposals~\cite{alexe10cvpr,uijlings13ijcv,dollar14eccv} (instances).
The task is both to localize the objects (select the best proposal) in the training images and to learn a detector for the target class.
To improve MIL, we transfer knowledge from $\aset$ to $\bset$, incorporating new cues into it.

Fig.~\ref{Fig:Overview} illustrates our transfer.
We first acquire three types of knowledge in the source $\aset$ (Sec.~\ref{Sec:acquire}):
1) a semantic segmentation model (Sec.~\ref{Sec:Segmentation model}),
2) the thing class similarities between $\aset$ and $\bset$ (Sec.~\ref{Sec:Similarity relation}) and
3) the co-occurrence frequencies between thing and stuff classes in $\aset$ (Sec.~\ref{Sec:Co-occurrence relation}).
Afterwards, we transfer the knowledge to $\bset$ (Sec.~\ref{Sec:transfer}).
Given an image in $\bset$, we first use the segmentation model to generate the thing ($T$) and stuff ($S$) maps of it (Sec.~\ref{Sec:generatemap}).
$T$ contains one score map ($R$) and one label ($L$) map, so does $S$. The segmentation model transfers knowledge generically to every image in $\bset$.
Building upon its result, we propose three proposal scoring schemes: label weighting (LW, Sec.~\ref{Sec:LW}), contrast weighting (CW, Sec.~\ref{Sec:CW}), and area weighting (AW, Sec.~\ref{Sec:AW}).
These link the pixel level segmentation to the proposal level score.
In each scheme, two scoring functions are proposed separately on thing and stuff maps.
We combine the three schemes to provide an even better proposal score to help MIL (Sec~\ref{Sec:SSI}).

\para{Scoring schemes.}
LW transfers the similarity and co-occurrence relations as weighting functions to the thing and stuff label maps, respectively.
Since we do not have stuff annotations on $\bset$, we conduct the co-occurrence knowledge transfer as a second-order transfer by finding the target class' most similar thing class in $\aset$.
We believe that the target class should appear against a similar background with its most similar class.
For example, in Fig.~\ref{Fig:Overview} target class bear's most similar class in $\aset$ is cat, LW up-weights the cat score on $T$ and its frequently co-occurring tree score on $S$.

LW favours small proposals with high weighted scores.
To counter this effect, we introduce the CW score.
It measures the dissimilarity of a proposal to its surroundings, measured on the thing/stuff score maps (Fig.~\ref{Fig:MC}).
CW up-weights proposals that are more likely to contain an entire object in $T$ or an entire stuff region in $S$.

Finally, the AW score encourages proposals to incorporate as much as possible of the connected components of pixels on a target's $K$ most similar classes in $\aset$ (\eg Fig.~\ref{Fig:Overview}: the cat area in the $T$ map).
While CW favors objects in general, AW focuses on objects of the target class in particular.

\section{Acquiring knowledge from the source $\aset$}\label{Sec:acquire}

\subsection{Segmentation model}\label{Sec:Segmentation model}
We employ the popular fully convolutional network (FCN-16s)~\cite{long15cvpr} to train an end-to-end semantic segmentation model on both thing and stuff classes of $\aset$.
Given a new image, the FCN model is able to predict a likelihood distribution over all classes at each pixel. Notice that the FCN model is first pretrained for image classification on ILSVRC 2012~\cite{russakovsky15ijcv}, then fine-tuned for semantic segmentation on $\aset$.
While it is possible that some of the target classes are seen during pretraining, only image-level labels are used. Therefore the weakly supervised setting still holds for the target classes.

\subsection{Similarity relations}\label{Sec:Similarity relation}
We compute the thing class similarities $V(a^{t}, b^t)$ between any thing class pair $(a^{t}, b^t)$.
We propose two similarity measures to compute $V$ as follows:

\para{Appearance similarity.}
Every image in $\aset$ or $\bset$ is represented by a 4096-dimensional CNN feature vector covering the whole image, using the output of the fc7 layer of the AlexNet CNN architecture~\cite{krizhevsky12nips}.
The similarity of two images is the inner product of their feature vectors. The similarity $V_{\mathrm{APP}}(a^{t}, b^t)$
is therefore the average similarity between images in $a^t$ and images in $b^t$.

\para{Semantic similarity.}
We compute the commonly used Lin~\cite{lin98icml} similarity $V_{\mathrm{SEM}}(a^{t}, b^t)$ between two nouns $b^t$ and $a^{t}$ in the WordNet hierarchy~\cite{fellbaum1998mit}.

\subsection{Co-occurrence relation}\label{Sec:Co-occurrence relation}
We denote by $U(a^{s},a^{t})$ the co-occurrence frequency of any stuff and thing class pair $(a^{s},a^{t})$ in $\aset$.
This frequency is computed and normalized over all the images in $\aset$.

\section{Transferring knowledge to the target $\bset$}\label{Sec:transfer}
This section transfers the source knowledge to the target set $\bset$.
In this set, we have access only to image level labels, but no location annotations.
We call the classes that are listed on the image level label list \emph{target classes}.
Given a new image of class $b^t$, we first use the FCN model trained on $\aset$ to generate the thing ($T$) and stuff ($S$) segmentations separately (Sec.~\ref{Sec:generatemap}).
Then we introduce three proposal scoring schemes to propagate the information from pixel level to proposal level (Sec.~\ref{Sec:LW} - \ref{Sec:AW}).
Finally we combine the three scoring schemes into a single window score (Sec.~\ref{Sec:SSI}). The scoring scheme parameters are learned in Sec.~\ref{sec:parameter}.

\subsection{Generating thing and stuff segmentations}\label{Sec:generatemap}
We apply the trained FCN model (Sec.~\ref{Sec:Segmentation model}) to a target image in $\bset$.
Usually, the output semantic segmentation is obtained by maximizing over all the class scores at each pixel~\cite{long15cvpr, chen15iclr, eigen15iccv, farabet13pami, noh15iccv, pinheiro14icml}.
In this paper, we instead generate two output segmentations, one for things $T$ and one for stuff $S$. We denote $i$ as the $i$-th pixel in the image.
We use $R^T = \{r^T_{i}\}$ and $L^T = \{l^T_{i}\}$ to denote the score ($R$) and label ($L$) maps for $T$.
They are generated by keeping the maximum score and the corresponding label over all the thing classes $\aset^T$ at each pixel $i$.
Similar to $R^T$ and $L^T$, $R^S = \{r^S_i\}$ and $L^S = \{l^S_i\}$ are generated by keeping the maximum score over all the stuff classes $\aset^S$ at each pixel.

Fig.~\ref{Fig:Overview} shows an example of a bear image (target). The thing and stuff maps are produced by the semantic segmentation model.
The $R$ heatmaps indicate the probability of assigning a certain thing or stuff label to each pixel.
Building upon these heatmaps, we propose three proposal scoring schemes to link the pixel level result to the proposal level score (Sec.~\ref{Sec:LW} - \ref{Sec:AW}).
These try to give high scores to proposals containing the target class.

\subsection{Label weighting (LW)}\label{Sec:LW}
Because bear is more similar to cat than to table, we want to up-weight the proposal area in the thing map if it is predicted as cat.
Meanwhile, because bear frequently appears against tree, we also want to up-weight the proposal area in the stuff map if it is predicted as tree.
To do this, we transfer the knowledge of similarity and co-occurrence relations acquired in the source to the target class (bear),
and use both relations to modulate the segmentation scores in $T$ and $S$.
Both relations and segmentation scores play a role in the label weighting proposal scoring scheme.

\para{Thing label weighting.}
We can generate a thing label weighting map depending on how close the predicted class $l^T_{i}$ at pixel $i$ in $L^T$ is to the target class $b^t$.
The thing label ($l^T_{i}$) weight is given by the class similarity score $V(l^T_{i},b^t)$ (Sec.~\ref{Sec:Similarity relation}).
In Fig.~\ref{Fig:Overview} the target class bear is more similar to cat than to table.
If a pixel is predicted as cat, then we assign a high label weight, otherwise we assign a low one.

\begin{figure}
	\centering
	\includegraphics[width=1\columnwidth]{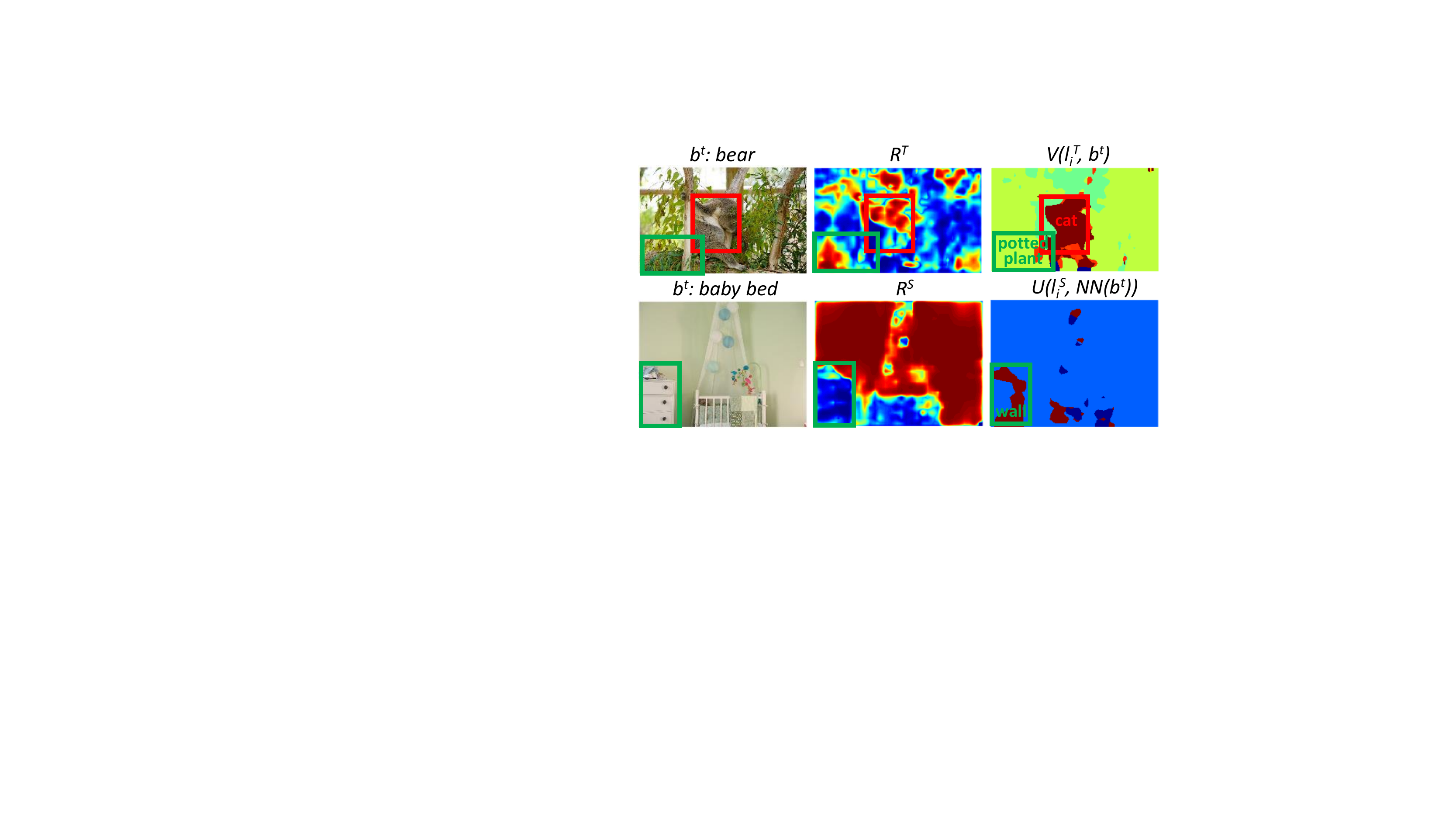}
         \vspace{-2mm}
         \prefigspace
	\caption{\small Label weighting example. Top: thing label weighting (class bear); bottom: stuff label weighting (class baby bed). $R^T$ and $R^S$ denote the thing and stuff score heatmaps, respectively; while $V(l^T_i,b^t)$ and $U(l^S_i,\mathrm{NN}(b^t))$ denote the thing and stuff label weighting heatmaps. We illustrate
some proposals in each image. We print the dominantly predicted labels in the proposals to show how label weighting favours $b^t$'s
NN class in thing maps and its frequently co-occurring stuff class
in stuff maps.
}
	\label{Fig:CT}
         \postfigspace
\end{figure}

\para{Stuff label weighting.}
We do not have stuff annotations in $\bset$. To conduct the stuff label weighting, we first find $b^t$'s most similar thing class in $\aset^T$ according to a similarity relation $V$ (we denote it by $\mathrm {NN}(b^t)$).
We believe that $b^t$ should
appear against a similar background (stuff) as its most similar thing class $\mathrm {NN}(b^t)$. We employ the co-occurrence frequency $U(l^S_i, \mathrm {NN}(b^t))$ of $\mathrm {NN}(b^t)$ as the corresponding stuff label weight for $l^S_{i}$ at pixel $i$ as stuff label weighting $L^S$.

In Fig.~\ref{Fig:Overview}, cat frequently co-occurs with trees, and so does bear. So, if a certain pixel is predicted as tree, it gets assigned a high stuff label weight.

\para{Proposal scoring.}
To score the proposals in an image, we multiply the label weights $V(l^T_{i},b^t)$ and $U(l^S_{i}, \mathrm {NN}(b^t))$ with the segmentation scores $r^T_{i}$ and $r^S_{i}$ at each pixel.
The weighting scheme is conducted separately on $T$ and $S$. Given a window proposal $w$, we average the weighted scores inside $w$:
\begin{equation}\label{Eqn:Wobj}
\begin{array}{*{20}{l}}
    \mathrm{LW}^{t}(w, \alpha^t) = f(\frac{1}{|w|}\sum\nolimits_{i \in w} {r^T_{i}V(l^T_{i},b^t)}, \alpha^t) \\
    \mathrm{LW}^{s}(w, \alpha^s) = f( \frac{1}{|w|}\sum\nolimits_{i \in w} {r^S_{i}U(l^S_{i}, \mathrm {NN}(b^t))}, \alpha^s)
\end{array}
\end{equation}
where $|w|$ denotes the size of $w$ (area in pixels).
We apply an exponential function $f(x) = \mathrm{exp}(\alpha \cdot x)$ to both thing and stuff LWs, $\alpha^t$ and $\alpha^s$ are the parameters.

Fig.~\ref{Fig:CT} offers two examples (bear and baby bed) for our thing and stuff label weighting schemes.
The red proposal in the top row is mostly classified as a cat and the green proposal as a potted plant.
Both proposals have high scores in the thing score map $R^T$, but the red proposal has a higher thing label weight $V(l^T_i,b^t)$, because cat is more similar to bear than to potted plant.
In contrast, the green proposal in the bottom row has low scores in $R^S$ but a high label weight $U(l^T_i,\mathrm{NN}(b^t))$, as baby bed co-occurs more frequently with wall.

Notice that the thing label weighting can be viewed as a first-order transfer where the information goes directly from the source thing classes to the target thing classes.
Instead, the stuff label weighting can be viewed as second-order transfer where the information first goes from the source stuff classes to the source thing classes, and then to the target thing classes.
To the best of our knowledge, such second-order transfer has not been proposed before.

\subsection{Contrast weighting (CW)}\label{Sec:CW}
The LW scheme favours small proposals with high label weights, which typically cover only part of an object (top right image in Fig.~\ref{Fig:Overview}). To counter this effect, contrast weighting (CW) measures the dissimilarity of a proposal to its immediate surrounding area on the thing/stuff score maps. It up-weights proposals that are more likely to contain an entire object or an entire stuff region.

\begin{figure}
	\centering
	\includegraphics[width=0.8\columnwidth]{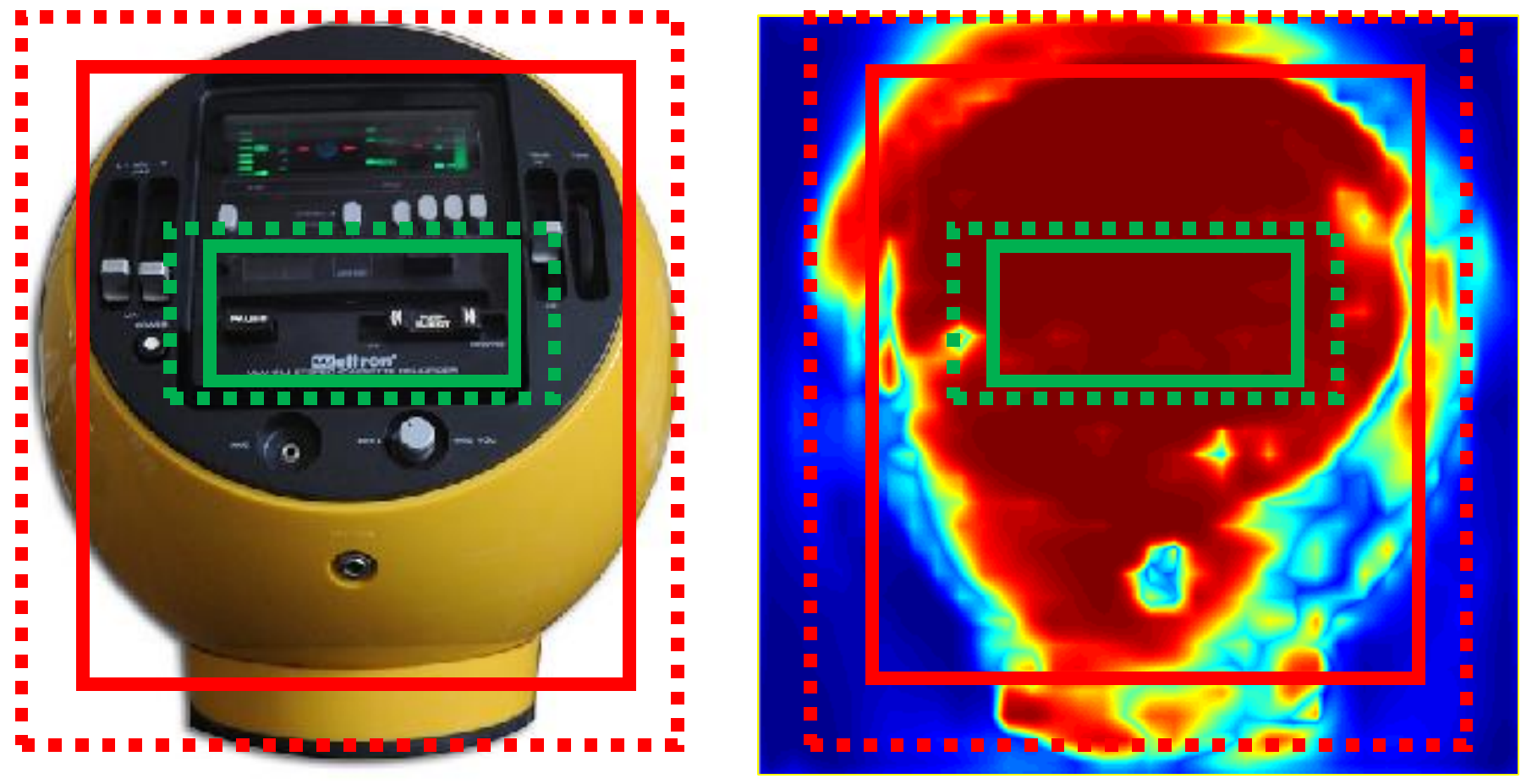}
         \prefigspace
	\caption{\small Contrast weighting example. An image of a tape player and some of its window proposals (left). $\mathrm{CW}^{t}$ is computed on the thing score map $R^t$ (right). The red proposal has a higher contrast $\mathrm{CW}^{t}$ (with its surrounding dashed ring) than the green one.
}
	\label{Fig:MC}
         \postfigspace
\end{figure}

The surrounding $Surr(w,\theta)$ of a proposal $w$ is a rectangular ring obtained by enlarging it by a factor $\theta$ in all directions~\cite{alexe10cvpr} (Fig.~\ref{Fig:MC}, the yellow ring).
The CW between a window and its surrounding ring is computed as the Chi-square distance between their score map ($R$) histograms $h(\cdot)$
\begin{equation}\label{Eqn:CW}
\mathrm{CW}(w, \theta) = {\chi ^2}(h(w),h(Surr(w,\theta )))
\end{equation}
We apply the CW scheme on both $R^T$ and $R^S$ and obtain $\mathrm{CW}^{t}(w, \theta^{t})$ and $\mathrm{CW}^{s}(w,\theta^{s})$. In Fig.~\ref{Fig:MC} the red proposal has a higher $\mathrm{CW}^{t}$ score compared to the green one.

\subsection{Area weighting (AW)}\label{Sec:AW}

\begin{figure}
	\centering
	\includegraphics[width=1\columnwidth]{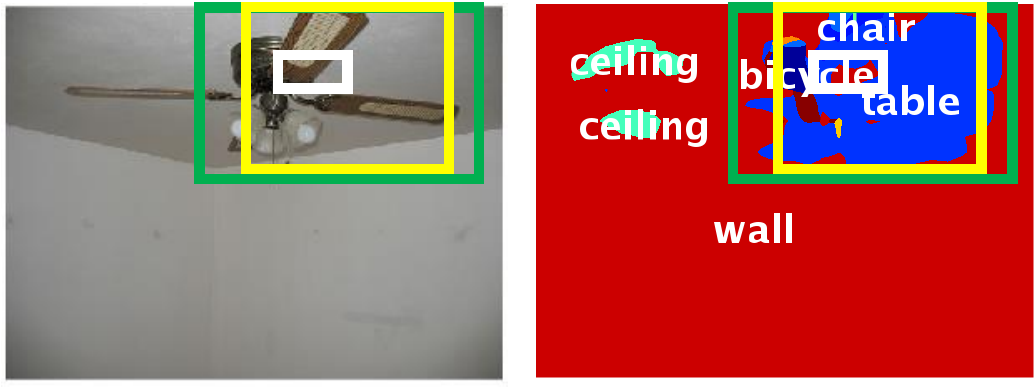}
         \vspace{-2mm}
         \prefigspace
	\caption{\small Area weighting example. An image of an electric fan (left) and its semantic segmentation (right). Thing area weighting favours the yellow proposal compared to the white one, as it incorporates most of the the connected component area of table and chair. Stuff area weighting further favours the green proposal as it allows certain stuff area in a proposal as the surrounding area of electric fan.}
	\label{Fig:PB}
         \postfigspace
\end{figure}

\para{Thing area weighting.}
Fig.~\ref{Fig:PB} gives an example of an electric fan and its semantic segmentation map.
Its $3\mathrm{\text{-}NN}$ classes in terms of appearance similarity (Sec.~\ref{Sec:Similarity relation}) are table, chair and people.
Between the white and yellow proposals, the CW scheme gives a bigger score to the white one, because its contrast is high.
Instead, the yellow proposal incorporates most of the fan area, but is unfortunately predicted as table and chair.
The thing area weighting scheme helps here boosting the yellow proposal's score.
We find the $K$-NN classes of $b^t$ in $\aset^T$ by using one of the similarity measures in Sec.~\ref{Sec:Similarity relation}.
Given a window $w$, we denote by $Area(w, b^t)$ the segment areas of any $K\mathrm{\text{-}NN}(b^t)$ inside $w$;
while $Area(O(w), b^t)$ is the area that expands the current segments to their connected components inside and outside $w$.
We measure the area ratio between the segments and their corresponding connected components:
\begin{equation}\label{Eqn:TAW}
    Ratio^t(w) = \frac{{Area(w, b^t)}}{{Area(O(w), b^t)}}
\end{equation}

If none of the $K$-NN classes occurs in $w$, we simply set $Ratio^t$ to zero.
Throughout this paper, $K$ is set to 3.

\para{Stuff area weighting.}
In Fig.~\ref{Fig:PB} among the three proposals, the green one is the best detection of the fan.
However, its score is not the highest according to $\mathrm{LW}^{t}$, $\mathrm{CW}^{t}$ and $\mathrm{AW}^{t}$, as it contains some stuff area (wall) surrounding the fan.
A bounding box usually has to incorporate some stuff area to fit an object tightly, as objects are rarely perfectly rectangle-shaped.
We propose to up-weight a window $w$ if stuff occupies a small but non-zero fraction of the window.
We denote with $Ratio^s(w)$ the percentage of stuff pixels in window $w$.

For thing and stuff area weighting we apply a cumulative distribution function (CDF) of the normal distribution
\begin{equation}\label{Eqn:CC2}
\begin{array}{*{20}{c}}
    \mathrm{AW}^{t}(w, \mu^t, \sigma^t) = \mathrm{CDF}(Ratio^{t}(w)| \, \mu^t, \sigma^t) \\
    \mathrm{AW}^{s}(w, \mu^s, \sigma^s) = \mathrm{CDF}(Ratio^{s}(w)| \, \mu^s, \sigma^s)
\end{array}
\end{equation}
where $\mu^t$ and $\sigma^t$ are the mean and standard deviation.
We choose $\mu^t = \mu^s = 0$ and $\sigma^{t}$, $\sigma^{s}$ are free parameters (Sec.~\ref{sec:parameter}).

\subsection{Combining the scoring schemes}
\label{Sec:SSI}
For each proposal in an image, the above scoring schemes can be independently computed, each on the thing and stuff map.
The scoring schemes tackle different problems, and are complementary to each other. This sections combines them to give our final TST (things and stuff transfer) window score $W$.

All the scoring functions on the thing map are
multiplied together as a thing score $W^t = \mathrm{LW}^t*\mathrm{CW}^t*\mathrm{AW}^t$.
This gives a higher score if a proposal mostly contains a target thing labeled as present in that image. Similarly, we have the stuff score
$W^s = \mathrm{LW}^s*\mathrm{CW}^s*\mathrm{AW}^s$, which gives a higher score if a proposal mostly contains stuff. To combine the thing and stuff scores, we simply subtract $W^s$ from $W^t$
\begin{equation}\label{Eqn:SSI}
\begin{array}{*{20}{c}}
W = W^t - W^s
\end{array}
\end{equation}

\subsection{Parameter learning}
\label{sec:parameter}

In the WSOL setting, we do not have the ground truth bounding box annotations in the target set $\bset$.
Thus we learn the score parameters $\alpha^t$, $\alpha^s$, $\theta^t$, $\theta^s$, $\sigma^t$ and $\sigma^s$ on the source set $\aset$, where we have ground truth.
We train the semantic segmentation model on the \emph{train} set of $\aset$, and then apply it to the \emph{val} set of $\aset$.
For each image in the \emph{val} set, we rank all its proposals using (\ref{Eqn:SSI}).
We jointly learn the score parameters by maximizing the performance over the entire validation set.

\section{Overall system}
\label{Sec: Eval}

In WSOL, given the target training set in $\bset$ with image level labels, the goal is to localize the object instances in it and to train good object detectors for the target test set.
We explain here how we build a complete WSOL system by building on a MIL framework and incorporating our transfer cues into it.

\para{Basic MIL.}
We build a Basic MIL pipeline as follows.
We represent each image in the target set $\bset$ as a bag of object proposals extracted
using Edge Boxes~\cite{dollar14eccv}.
They return about 5,000 proposals per image, likely to cover all objects.
Following~\cite{girshick14cvpr,bilen14bmvc,song14icml,song14nips,wang15tip}, we describe the proposals by the output of the fc7 layer of the AlexNet CNN architecture~\cite{krizhevsky12nips}.
The CNN model is pre-trained for whole-image classification on ILSVRC~\cite{russakovsky15ijcv}, using the Caffe implementation~\cite{jia13caffe}. This produces a 4,096-dimensional feature vector for each proposal. Based on this
feature representation for each target class, we iteratively build an SVM appearance model (object detector) in two alternating steps:
(1) Re-localization: in each positive image, we select the highest scoring proposal by the SVM. This produces the positive set which contains the current selection of one instance from each positive image.
(2) Re-training: we train the SVM using the current selection of positive samples, and all proposals from the negative images as negative samples.
As in~\cite{deselaers10eccv,siva11iccv,cinbis15pami,guillaumin12cvpr,tang14cvpr}, we also linearly combine the SVM score with a general measure of objectness~\cite{alexe10cvpr,dollar14eccv}. This leads to a higher MIL baseline.

\para{Incorporating things and stuff transfer (TST).}
We incorporate our things and stuff transfer (TST) into Basic MIL by linearly combining the SVM score with our proposal scoring function (\ref{Eqn:SSI}).
Note how the behavior of (\ref{Eqn:SSI}) depends on the class similarity measure used within it (either appearance or semantic similarity, Sec.~\ref{Sec:Similarity relation}).

\para{Deep MIL.}
Basic MIL uses an SVM on top of fixed deep features as the appearance model. Now we change the model to fine-tune all layers of the deep network during the re-training step of MIL. We take the output of Basic MIL as an initialization for two additional MIL iterations. During these iterations, we use Fast R-CNN~\cite{girshick15cvpr}.

\section{Experiments}\label{sec:experiment}

\begin{table}
	\setlength{\tabcolsep}{2.6pt}
	\centering
	\small
\begin{tabular}{|l|c|c|}
          \hline
	 Method		&	APP	&	SEM	\\
          \hline
	 \hline
          Basic MIL	&	\multicolumn{2}{c|}{39.7} 	\\
          \hline
           DT $\approx$~\cite{rochan15cvpr} (transfer only) 	&	15.0		&	 -			\\ 
           DT + MIL $\approx$~\cite{rochan15cvpr} (full) 	&	39.5		&	 -			\\ 
            TST	&	46.7		&	 46.0			\\
          \hline
          \hline
          Basic MIL + Objectness~\cite{dollar14eccv}	&	\multicolumn{2}{c|}{47.6} 		\\
         \hline
           DT	$\approx$~\cite{rochan15cvpr} (transfer only) &	34.5		&	 -			\\ 
           DT + MIL $\approx$~\cite{rochan15cvpr} (full)	&	49.1		&	 -			\\ 
           TST	&	52.7		&	 52.5			\\
	 \hline
\hline
   Deep MIL + Objectness~\cite{dollar14eccv}	&	\multicolumn{2}{c|}{48.4} 		\\
         \hline
           TST	&	54.0		&	 53.8			\\
           TST + ILSVRC-dets &	-	&	 \textbf{55.1}		\\
           \hline
\end{tabular}
         \pretabspace
	\caption{\small
CorLoc on ILSVRC-20; DT: direct transfer; DT+MIL: direct transfer plus MIL. TST is our method; ILSVRC-dets: Sec.~\ref{Sec:WSOL}, last paragraph.
The transfers are guided by either the semantic (SEM) or the appearance (APP) class similarity.
	}
	\label{Tab:Method}
         \posttabspace
\end{table}

\begin{table*}
         \vspace{-2mm}
	\setlength{\tabcolsep}{1.35pt}
	\centering
	\small
	\begin{tabular}{|c|cccccccccccc|cccccccc|cc|} \hline
		Class	& ant & bbed&bask&bear&burr&butt&cell&cmak&efan& elep&gfis&gcar&monk&pizz&rabb&stra&tpla&turt&wiro&whal& Avg. & Avg.(8)\\
                   \hline
		\hline
                   LSDA~\cite{hoffman14nips} & - & - & - & - & - & - & - & - & - & - & - & - & 22.9 & 27.6 & 40.2 & 6.8 & 19.1 & 31.9 & 8.6 & \textbf{20.3} & - & 22.2 \\	
		\hline
		Deep MIL+Obj.&39.2&	24.2&	0.2&	13.0&	16.5&	28.9&	29.7&	 8.9&	\textbf{39.1}&	34.4&	9.1&	40.3&	18.0&	29.7&	 32.8&	19.6&	 27.0&	27.0&	5.9&	2.9&	22.3 & 20.4		\\
		+TST (APP)&\textbf{39.9}&	\textbf{31.0}&	0.6&	16.8&	 11.3&	32.2&	32.0&	6.0&	34.9&	38.4&	 \textbf{13.6}&	 \textbf{65.1}&	23.8&32.5&	 40.7&	 \textbf{24.8}&	28.6&	25.1&	\textbf{9.9} &	5.1&25.6 & 23.8 \\
		+TST (SEM)&34.1&	26.8&	0.6&19.7&	16.8&	31.7&	 \textbf{32.6}&	8.6&	31.2&	37.2&	11.5&	57.8&	22.9&	31.2&	 \textbf{45.2}&	18.7&	 \textbf{30.3}&	28.2 &	8.1&	 6.2&25.0	& 23.9 \\	
+ ILSVRC-dets &34.1&	24.7&	\textbf{3.3}&\textbf{21.5}&	\textbf{18.6}&	\textbf{35.1}&	 \textbf{32.6}&	\textbf{9.1}&	32.9&	\textbf{38.8}&	11.1&	58.5&	 \textbf{24.5}&	\textbf{33.9}&	 44.5&	18.4&28.4&	\textbf{32.1} &	9.9&	 5.7&\textbf{25.9}	& \textbf{24.7} \\	
		\hline
	\end{tabular}
         \pretabspace
	\caption{\small mAP Performance on the test set of ILSVRC-20. All our methods start from DeepMIL with objectness.
For comparison we also show the performance on the 8 classes common to our target set and that of LSDA~\cite{hoffman14nips}.}\label{Tab:mAP}
         \posttabspace
\end{table*}

\subsection{Datasets and evaluation protocol}
\label{sec:dataset}

We use one source set $\aset$ (PASCAL Context) and several different target sets $\bset$ in turn (ILSVRC-20, COCO-07 and PASCAL VOC 2007).
Each target set contains a training set and a test set. We perform WSOL on the target training set to localize objects within it. Then we train a Fast R-CNN~\cite{girshick15iccv} detector from it and apply it on the target test set.

\para{Evaluation protocol.}
We quantify localization performance in the target training set with the CorLoc measure~\cite{bilen15cvpr,cinbis15pami,deselaers10eccv,shi15pami,wang15tip,bilen16cvpr}.
We quantify object detection performance on the target test set using mean average precision (mAP).
As in most previous WSOL methods~\cite{bilen14bmvc,bilen15cvpr,cinbis14cvpr,cinbis15pami,deselaers10eccv,russakovsky12eccv,siva11iccv,song14icml,song14nips,wang15tip}, our scheme returns exactly one bounding-box per class per training image.
At test time the object detector is capable of localizing multiple objects of the same class in the same image (and this is captured in the mAP measure).

\para{Source set: PASCAL Context.}
PASCAL Context \cite{mottaghi14cvpr} augments PASCAL VOC 2010~\cite{everingham10ijcv} with class labels at every pixel.
As in~\cite{mottaghi14cvpr}, we select the 59 most frequent classes.
We categorize them into \emph{things} and \emph{stuff}.
There are 40 thing classes, including the original 20 PASCAL classes and new classes such as book, cup and window.
There are 19 stuff classes, such as sky, water and grass.
%
%
We train the semantic segmentation model (Sec.~\ref{Sec:Segmentation model}) on the \emph{train} set of $\aset$ and set the score parameters (Sec.~\ref{sec:parameter}) on the \emph{val} set, using the 20 PASCAL classes from $\aset$ as targets.

\para{Target set: ILSVRC-20.}
The ILSVRC~\cite{russakovsky15ijcv} dataset originates from the ImageNet dataset~\cite{deng09cvpr}, but is much harder~\cite{russakovsky15ijcv}.
As the target training set we use the \emph{train60k} subset~\cite{girshick14cvpr} of ILSVRC 2014.
As the target test set we use the 20k images of the validation set.
To conduct WSOL on \emph{train60k}, we carefully select 20 target classes: ant, baby-bed, basketball, bear, burrito, butterfly, cello, coffee-maker, electric-fan, elephant, goldfish, golfcart, monkey, pizza, rabbit, strainer, tape-player, turtle, waffle-iron and whale.
ILSVRC-20 contains 3,843 target training set images and 877 target test set images.
This selection is good because:
(1) they are visually considerably different from any source class;
(2) they appear against similar background classes as the source classes, so we can show the benefits of stuff transfer;
(3) they are diverse, covering a broad range of object types.

\para{Target set: COCO-07.}
The COCO 2014~\cite{lin14eccv} dataset has fewer object classes (80) than ILSVRC (200), but more instances.
COCO is generally more difficult than ILSVRC for detection, as objects are smaller~\cite{lin14eccv}.
There are also more instances per image: 7.7 in COCO compared to 3.0 in ILSVRC~\cite{lin14eccv}.
We select 7 target classes to carry out WSOL: apple, giraffe, kite, microwave, snowboard, tennis racket and toilet.
COCO-07 contains 11,489 target training set images and 5,443 target test set images.

\para{Target set: PASCAL VOC 2007.}
The PASCAL VOC 2007~\cite{everingham15ijcv} dataset is one of the most important object detection datasets.
It includes 5,011 training (\emph{trainval}) images and 4,952 test images, which we directly use as our target training set and target test set, respectively.
For our experiments we use all 20 thing classes in VOC 2007.
Since the thing classes in our source set (PASCAL Context) overlap with those of VOC 2007, when doing our TST transfer to a target class we remove it from the sources.
For example, when we transfer to ``dog" in VOC 2007, we remove ``dog" from the FCN model trained on PASCAL Context.

\subsection{ILSVRC-20}\label{Sec:WSOL}
Table~\ref{Tab:Method} presents results for our method (TST) and several alternative methods on ILSVRC-20.

\begin{figure*}
	\centering
	\includegraphics[width=0.99\textwidth]{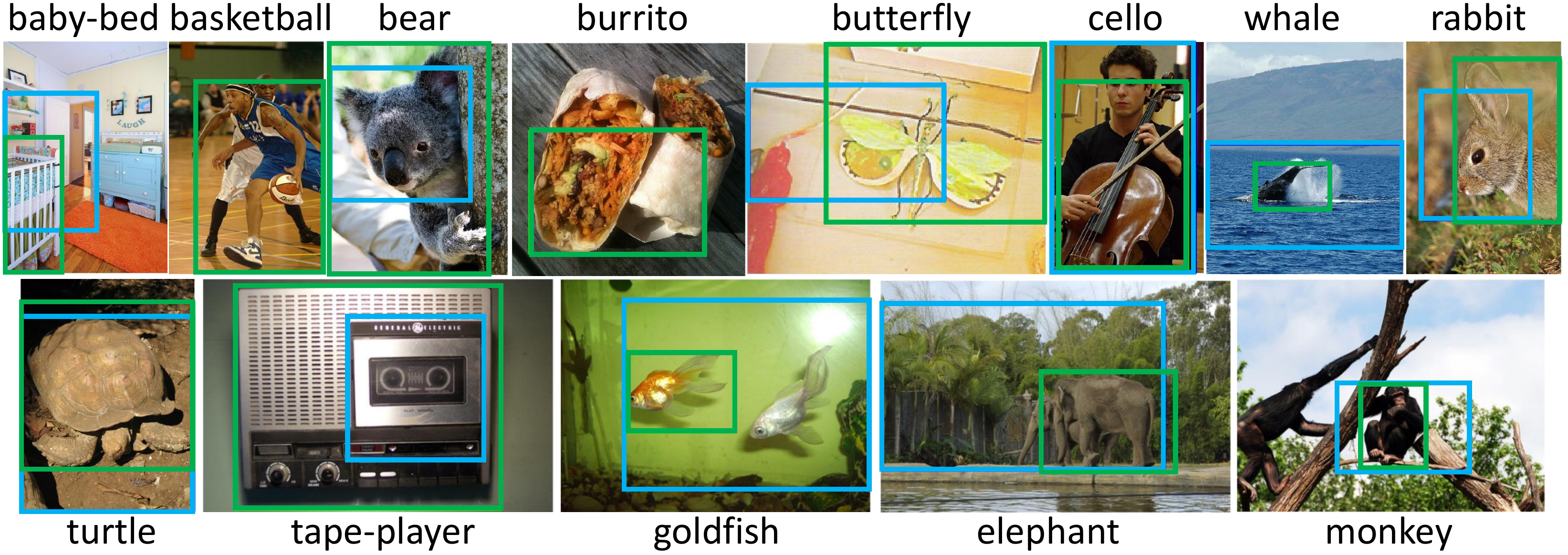}
         \prefigspace
         \vspace{0.5mm}
	\caption{\small We show localizations on ILSVRC-20 of Basic MIL with objectness (blue) and our TST (APP) method (green).
}
	\label{Fig:Result-Figure-same}
         \postfigspace
\end{figure*}

\para{Our transfer (TST).}
Our results (TST) vary depending on the underlying class similarity measure used, either appearance (APP) or semantic (SEM) (Sec.~\ref{Sec:Similarity relation}).
TST (APP) leads to slightly better results than TST (SEM).
We achieve a $+7\%$ improvement in CorLoc (46.7) compared to Basic MIL without objectness, and $+5\%$ improvement (52.7) over Basic MIL with objectness.
Hence, our transfer method is effective, and is complementary to objectness.
Fig.~\ref{Fig:Result-Figure-same} shows example localizations by Basic MIL with objectness and TST (APP).

\para{Comparison to direct transfer (DT).}
We compare here to a simpler way to transfer knowledge.
We train a fully supervised object detector for each source thing class.
Then, for every target class we find the most similar source class from the 40 PASCAL Context thing classes, and use it to directly detect the target objects.
For the appearance similarity measure (APP) all NN classes of ILSVRC-20 are part of PASCAL VOC and PASCAL Context.
Therefore we have bounding box annotations for these classes.
However, for the semantic similarity measure (SEM) not all NN classes of ILSVRC-20 are part of PASCAL VOC.
Therefore we do not have bounding box annotations for these classes and cannot apply DT.
DT is similar to the `transfer only' method in~\cite{rochan15cvpr} (see Sec. 4.2 and Table 2 in~\cite{rochan15cvpr}).

As Table~\ref{Tab:Method} shows, the results are quite poor as the source and target classes are visually quite different,
\eg the most similar class to ant according to APP is bird; while for waffle-iron, it is table; for golfcart, it is person.
This shows that the transfer task we address (from PASCAL Context to ILSVRC-20) is challenging and cannot be solved by simply using object detectors pre-trained on the source classes.

\para{Comparison to direct transfer with MIL (DT+MIL).}
We improve the direct transfer method by using the DT detector to score all proposals in a target image, and then combining this score with the standard SVM score for the target class during the MIL re-localization step.
This is very similar to the full method of~\cite{rochan15cvpr} and is also close to~\cite{guillaumin12cvpr}.
The main difference from~\cite{rochan15cvpr} is that we train the target class' SVM model in an MIL framework (Sec.~\ref{Sec: Eval}), whereas~\cite{rochan15cvpr} simply trains it by using proposals with high objectness as positive samples.

As Table~\ref{Tab:Method} shows, DT+MIL performs substantially better than DT alone, but it only slightly exceeds MIL without transfer, again due to the source and target classes being visually different ($+1.5\%$ over Basic MIL with objectness).
Importantly, our method (TST) achieves higher results, demonstrating that it is a better way to transfer knowledge ($+5\%$ over Basic MIL with objectness).

\para{Deep MIL.}
As Table~\ref{Tab:Method} shows, Deep MIL improves slightly over Basic MIL (from 47.6 to 48.4, both with objectness).
When built on Deep MIL, our TST transfer raises CorLoc to 54.0 (APP) and 53.8 (SEM), a $+5\%$ improvement over Deep MIL (confirming what we observed when building on Basic MIL).
Table~\ref{Tab:mAP} shows the mAP of Deep MIL and our method (TST) on the test set.
The observed improvements in CorLoc on the training set nicely translate to better mAP on the test set ($+3.3\%$ over Deep MIL).

\para{Comparison to LSDA~\cite{hoffman14nips}.}
We compare to LSDA~\cite{hoffman14nips}, which trains fully supervised detectors for 100 classes of the ILSVRC 2013 dataset (sources) and transfers to the other 100 classes (targets).
We report in Table~\ref{Tab:mAP} the mAP on the 8 classes common to both their target set and ours.
On these 8 classes, we improve on~\cite{hoffman14nips} by $+1.7\%$ mAP while using a substantially smaller source set (5K images in PASCAL Context, compared to 105K images in their 100 source classes from ILSVRC 2013).

Furthermore, we can also incorporate detectors for their 100 source classes in our method,
in a similar manner as for the DT+MIL method.
For each target class we use the detector of the $3$ most similar source classes as a proposal scoring function during MIL's re-localization step.
We choose the SEM measure to guide the transfer as it is fast to compute.
This new scoring function is referred to as ILSVRC-dets in Table~\ref{Tab:Method} and~\ref{Tab:mAP}.
When using the ILSVRC-dets score, our mAP improves further, to a final value $+2.5\%$ better than LSDA~\cite{hoffman14nips}.

\subsection{COCO-07}
\label{Sec:COCOres}

Table~\ref{Tab:COCO} presents results on COCO-07, which is a harder dataset.
Compared to Deep MIL with objectness, our transfer method improves CorLoc by $+3.0\%$ and mAP by $+2.2\%$ (APP).

\begin{table}
\small
	\setlength{\tabcolsep}{2.6pt}
	\centering
	\small
\begin{tabular}{|c|c|c|}
          \hline
	 Method		&	training (CorLoc)	&	test (mAP)	\\
          \hline
	 \hline
          Deep MIL + Obj.	&	15.8				&	9.1 			\\
          +TST (SEM)	&	18.0				&	11.0			\\
          +TST (APP)	&	\textbf{18.8}		&	\textbf{11.3}	\\
	 \hline
\end{tabular}
         \pretabspace
	\caption{\small
	CorLoc and mAP on COCO-07. Objectness~\cite{dollar14eccv} is added on top of the baseline. TST (SEM) and TST (APP) are separately added to the baseline with objectness.
	}
	\label{Tab:COCO}
         \posttabspace
\end{table}

\subsection{PASCAL VOC 2007}
\label{Sec:VOCres}

Table~\ref{Tab:VOC} presents results on PASCAL VOC 2007. As our baseline system, we use both objectness and multifolding~\cite{cinbis15pami} in Deep MIL. This performs at $50.7$ CorLoc and $28.1$ mAP.
Our transfer method TST strongly improves CorLoc to $59.9$ ($+9.2\%$) and mAP to $33.8$ ($+5.7\%$).

\para{Comparison to~\cite{rochan15cvpr}.}
They present results on this dataset in a transfer setting, by using detectors trained in a fully supervised setting for all 200 classes of ILSVRC (excluding the target class).
Adopting their protocol, we also use those detectors in our method (analog to the LSDA comparison above).
This leads to our highest CorLoc of $60.8$, which outperforms~\cite{rochan15cvpr}, as well as state-of-the-art WSOL works~\cite{wang15tip,bilen16cvpr,cinbis15pami} (which do not use such transfer).
For completeness, we also report the corresponding mAPs.
Our mAP 34.5 matches the result of~\cite{bilen16cvpr} based on their 'S' neural network, which corresponds to the AlexNet we use.
They propose an advanced WSOL technique that integrates both recognition and detection tasks to jointly train a weakly supervised deep network, whilst we build on a weaker MIL system.
We believe our contributions are complementary: we could incorporate our TST transfer cues into their WSOL technique and get even better results.

Finally, we note that our experimental protocol guarantees no overlap in either images nor classes between source and target sets (Sec.~\ref{sec:dataset}).
However, in general VOC 2007 and PASCAL Context (VOC 2010) share similar attributes, which makes this transfer task easier in our setting.

\begin{table}
\small
	\setlength{\tabcolsep}{1pt}
	\centering
	\small
\begin{tabular}{|c|c|c|c|} \hline
	Method		&   	ILSVRC-dets	&	CorLoc 		&	mAP	\\
	\hline
	\hline
         Wang et al.~\cite{wang15tip}	&	&	48.5					&	31.6			\\
         Bilen and Vedaldi~\cite{bilen16cvpr} (S)	&	&	54.2              			&	\textbf{34.5}	\\
         Cinbis et al.~\cite{cinbis15pami}	&	&	54.2					&	28.6			\\
         Rochan and Wang~\cite{rochan15cvpr}& \checkmark	& 	58.8					&	- 			\\
         \hline
	
 Deep MIL + Obj. + MF 				&			&	50.7					&	28.1	\\
+TST (SEM)				&		&	59.9					&	33.8 \\
+TST (SEM)				&	\checkmark		&	\textbf{60.8}					&	\textbf{34.5} \\
	\hline
\end{tabular}
         \pretabspace
	\caption{\small
         Performance on PASCAL VOC 2007. We start from Deep MIL with objectness~\cite{dollar14eccv} and multifolding~\cite{cinbis15pami} as a baseline. Then we add our method TST (SEM) to it. Rochan and Wang~\cite{rochan15cvpr} do not report mAP.
         (S) denotes the S model (roughly AlexNet) in~\cite{bilen16cvpr}, which corresponds to the network architecture we use in all experiments. ILSVRC-dets indicates using detectors trained from ILSVRC during transfer.}
	\label{Tab:VOC}
         \posttabspace
         \vspace{-3mm}
\end{table}

\section{Conclusion}\label{sec:conclusion}

We present weakly supervised object localization using things and stuff transfer.
We transfer knowledge by training a semantic segmentation model on the source set and using it to generate thing and stuff maps on a target image.
Class similarity and co-occurrence relations are also transferred and used as weighting functions.
We devise three proposal scoring schemes on both thing and stuff maps and combine them to produce our final TST score.
We plug the score into an MIL pipeline and show significant improvements on the ILSVRC-20, VOC 2007 and COCO-07 datasets.
We compare favourably to two previous transfer works~\cite{rochan15cvpr,hoffman14nips}. 
\textbf{Acknowledgements.} Work supported by the ERC Starting Grant VisCul. 

\title{Weakly Supervised Object Localization Using Things and Stuff Transfer\\
       --- Supplemental Material--- }

\author{
Miaojing Shi$^{1,2}$\\
\\
\and
Holger Caesar$^1$\\
\\
$^1$University of Edinburgh~~~$^2$Tencent Youtu Lab\\
{\tt\small name.surname@ed.ac.uk}
\and
Vittorio Ferrari$^1$\\
\\
}

\maketitle


\appendix
\begin{abstract}
The appendices provided in this supplemental material
complement our paper in two aspects. We provide the proxy measures to detail the parameter learning process of our method.
We conduct an ablation study to demonstrate the contribution of each component in the proposed system.
\end{abstract}

\section{Proxy measures}\label{sec:supplementary}
We propose two proxy measures to jointly learn the score parameters by maximizing the performance over the entire validation set in $\aset$ (Sec. 5.6):
\begin{enumerate}
\item Rank: the highest rank of any proposal whose intersection-over-union (IoU) with ground truth bounding box is $>$ 0.5.
\item CorLoc@1: the percentage of images in which the highest scoring proposal localizes an object of the target class correctly (IoU $>$ 0.5).
\end{enumerate}
These two measures characterize well whether a proposal scoring function gives a higher score to the target objects than to other proposals.
Hence they are good proxy measures for their usefulness within MIL. The behavior of the proposal scoring functions (Eqn. 5) depends on the class similarity measure used within them. Referring to Sec. 4.2, the guided similarities can be either appearance or semantic similarity (APP/SEM).

\begin{figure}
	\centering
	\includegraphics[width=1\columnwidth]{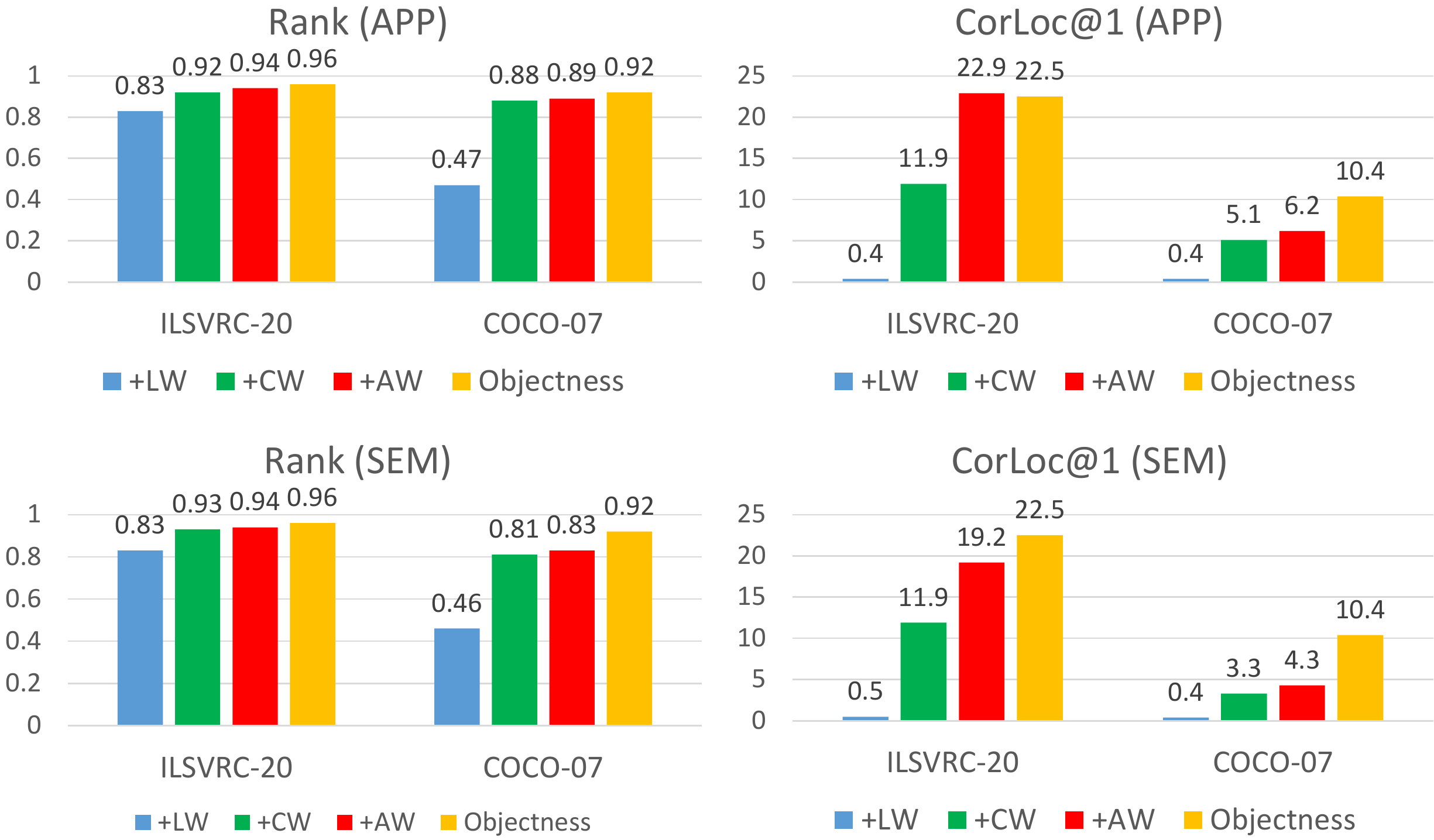}
	\caption{ Proxy measures on ILSVRC-20 and COCO-07. Our transfer is guided by the appearance similarity (top: APP) as well as the semantic similarity (bottom: SEM).
	Performance is measured by Rank (left) and CorLoc@1 (right) (see Sec. 5.6). }
	\label{Fig:ParaTest}
\end{figure}

\para{Results.}
We notice that roughly the same parameters are obtained from both criteria. Now we test how well they work on two of our target sets: ILSVRC-20 and COCO-07.
We gradually add each proposal scoring scheme from Sec.~5.2 - 5.4 and denote them by +LW, +CW, and +AW in Fig.~\ref{Fig:ParaTest}.
Both Rank and CorLoc@1 are gradually improved: using APP we achieve the highest CorLoc@1: 22.9 on ILSVRC-20 and 6.2 on COCO-07;
and the highest Rank: 0.94 on ILSVRC-20 and 0.89 on COCO-07.
SEM is lower than APP: 19.2 and 4.3 in terms of CorLoc@1, and 0.94 and 0.83 in terms of Rank, on ILSVRC-20 and COCO-07, respectively.
Comparing our proposal scoring schemes with a modern version of objectness~\cite{dollar14eccv},
we see that both perform similarly well.
In Sec.~7.2 and 7.3 we integrate our scheme with objectness and achieve a big improvement (+5\%), 
which shows that both are complementary.

\section{Ablation study}
We report here an ablation study to offer the justification of each component in our proposed system.
We incorporate the LW, AW and CW scores (Sec. 5.2 -5.4) separately into the Basic MIL framework (Sec. 7.1).
We report experiments on ILSVRC-20 in Table~\ref{Tab:Sup} in the same protocol as for the first table in the paper (guided by appearance similarity APP column).
The three scores bring +0.9\%, +3.3\%, and +3.2\% CorLoc on top of Basic MIL's 39.7\%.
This demonstrates that each individual score brings an improvement. Moreover, we also tried combining multiple scores: LW+CW reaches 44.0\%,
AW+CW reaches 45.8\%, and using all three scores AW+LW+CW gives us the highest CorLoc 47.6\%.
Here we can see that LW brings an additional improvement of +1.8\% when added to AW+CW. This shows that by carefully designing and integrating each component into our system, we are able to boost the overall performance over each individual component or any two of them.

\begin{table}
\small
	\centering
	\small
\begin{tabular}{|c|c|c|c|c|} \hline
	Method	& LW&	CW 		&	AW & CorLoc \\
	\hline
	\cline{1-5}
 \multirow{6}{*}{Basic MIL}    &  & & & 39.7   \\
 \cline{2-5}
  &  + & & &  40.6 \\
    &   & + & & 43.0 \\
      &  & &+ & 42.9 \\
     \cline{2-5}
        & + & + & & 44.0 \\
        & & + &+ & 45.8 \\
        \cline{2-5}
          & + & +&+ & 47.6 \\
	\hline
\end{tabular}
         \pretabspace
	\caption{Ablation study on ILSVRC-20. LW: label weighting; CW: contrast weighting; AW: area weighting. We start from Basic MIL and incorporate LW, CW, AW, or any of their combination into it. We report the CorLoc. }
	\label{Tab:Sup}
         \posttabspace
         \vspace{-3mm}
\end{table}

\clearpage
{\small
	\bibliographystyle{unsrt}
\bibliographystyle{ieee}
\bibliography{../../../../bibtex/shortstrings,../../../../bibtex/calvin,../../../../bibtex/vggroup}
}
\end{document}